\documentclass[conference]{IEEEtran}
\IEEEoverridecommandlockouts

\usepackage{cite}
\usepackage{amsmath,amssymb,amsfonts}
\usepackage{algorithmic}
\usepackage{graphicx}
\usepackage{textcomp}
\usepackage{xcolor}
\usepackage{amsmath}

\def\BibTeX{{\rm B\kern-.05em{\sc i\kern-.025em b}\kern-.08em
    T\kern-.1667em\lower.7ex\hbox{E}\kern-.125emX}}
\begin{document}

\title{DAF-Net: A Dual-Branch Feature Decomposition Fusion Network with Domain Adaptive for Infrared and Visible Image Fusion\\
}

\author{\IEEEauthorblockN{1\textsuperscript{st} Jian Xu}
	\IEEEauthorblockA{\textit{School of Information and Communication Engineering} \\
		\textit{University of Electronic Science and Technology of China}\\
		Chengdu, China \\
		xujian.0426@std.uestc.edu.cn}
	\and
	\IEEEauthorblockN{2\textsuperscript{nd} Xin He}
	\IEEEauthorblockA{\textit{School of Resources and Environment} \\
		\textit{University of Electronic Science and Technology of China}\\
		Chengdu, China \\
		hxadolf@163.com}
	\and
}

\maketitle

\begin{abstract}
	Infrared and visible image fusion aims to combine complementary information from both modalities to provide a more comprehensive scene understanding. However, due to the significant differences between the two modalities, preserving key features during the fusion process remains a challenge. To address this issue, we propose a dual-branch feature decomposition fusion network (DAF-Net) with domain adaptive, which introduces Multi-Kernel Maximum Mean Discrepancy (MK-MMD) into the base encoder and designs a hybrid kernel function suitable for infrared and visible image fusion. The base encoder built on the Restormer network captures global structural information while the detail encoder based on Invertible Neural Networks (INN) focuses on extracting detail texture information. By incorporating MK-MMD, the DAF-Net effectively aligns the latent feature spaces of visible and infrared images, thereby improving the quality of the fused images. Experimental results demonstrate that the proposed method outperforms existing techniques across multiple datasets, significantly enhancing both visual quality and fusion performance. The related Python code is available at https://github.com/xujian000/DAF-Net.
\end{abstract}

\begin{IEEEkeywords}
	Infrared and visible image fusion, Dual-branch network, Multi-Kernel Maximum Mean Discrepancy, Hybrid kernel function.
\end{IEEEkeywords}

\section{Introduction}
\label{sec:intro}
Infrared and visible image fusion combines complementary information from both modalities to provide a more comprehensive scene understanding \cite{ma2019infrared}. Infrared images excel at capturing thermal radiation, particularly in low-light or complex environments, such as night surveillance and target detection \cite{ma2019fusiongan}. Visible images retain rich details and color, offering clear scene representation. Fusing these modalities compensates for the limitations of each, achieving a more complete understanding of the environment. However, significant differences in imaging principles, resolution, and spectral response pose a challenge in maintaining the consistency of key features during fusion \cite{tang2022image}.

Existing image fusion methods can be broadly categorized into three types: traditional methods, transform-domain methods, and deep learning-based approaches. Traditional methods, such as pixel-level or feature-level fusion, rely on simple rules, making them computationally efficient and easy to implement. However, they often fail to fully exploit the complementary information between infrared and visible images, resulting in limited fusion performance \cite{ZHANG2021323}. While these methods are fast and easy to apply, they struggle to produce high-quality fused images that capture all details from both modalities. Transform-domain methods, such as wavelet transform \cite{pajares2004wavelet,hill2002image,nunez1999multiresolution} and Laplacian pyramid techniques \cite{wang2011multi,sahu2014medical,du2016union}, decompose images into different frequency components, preserving details to some extent. Despite their effectiveness in capturing multi-frequency details, key modality-specific features may be lost during reconstruction, making it difficult to retain both global structure and fine texture. Recently, deep learning-based methods have made significant strides. Techniques such as convolutional neural networks (CNNs) and generative adversarial networks (GANs) learn nonlinear relationships between modalities, achieving outstanding performance in image fusion \cite{10446402,pmgi,10448114,ma2020pan,ma2020ddcgan,10445881,10448503}. These methods generate fused images with higher visual quality by modeling modality interactions effectively. However, deep learning approaches typically require large amounts of labeled data, which can be a constraint when data is scarce \cite{ZHANG2021323}, and still face challenges in balancing the preservation of global structure and fine texture.

This paper proposes a domain-adaptive dual-branch feature decomposition fusion network (DAF-Net), introducing Multi-Kernel Maximum Mean Discrepancy (MK-MMD) \cite{borgwardt2006integrating} in the base encoder to better align latent features of infrared and visible images. The base encoder built on the Restormer network \cite{zamir2022restormer} captures global structural information and uses MK-MMD to reduce distributional differences at the global feature level. The detail encoder based on Invertible Neural Networks (INN) \cite{ardizzone2018analyzing} extracts detail texture information to preserve the unique characteristics of each modality. MK-MMD is applied only in the base encoder to ensure global feature consistency, avoiding over-alignment of local details and loss of modality-specific information. This structure enables DAF-Net to balance global structure and detail preservation. Experimental results show DAF-Net significantly improves visual quality and fusion performance across datasets.
\section{PROPOSED METHOD}
In this section, we introduce the network architecture of DAF-Net, followed by an introduction to the two-stage training process and loss functions.

\subsection{Network Architecture}
\begin{figure}[htb]
	\centering
	\includegraphics[width=8.5cm]{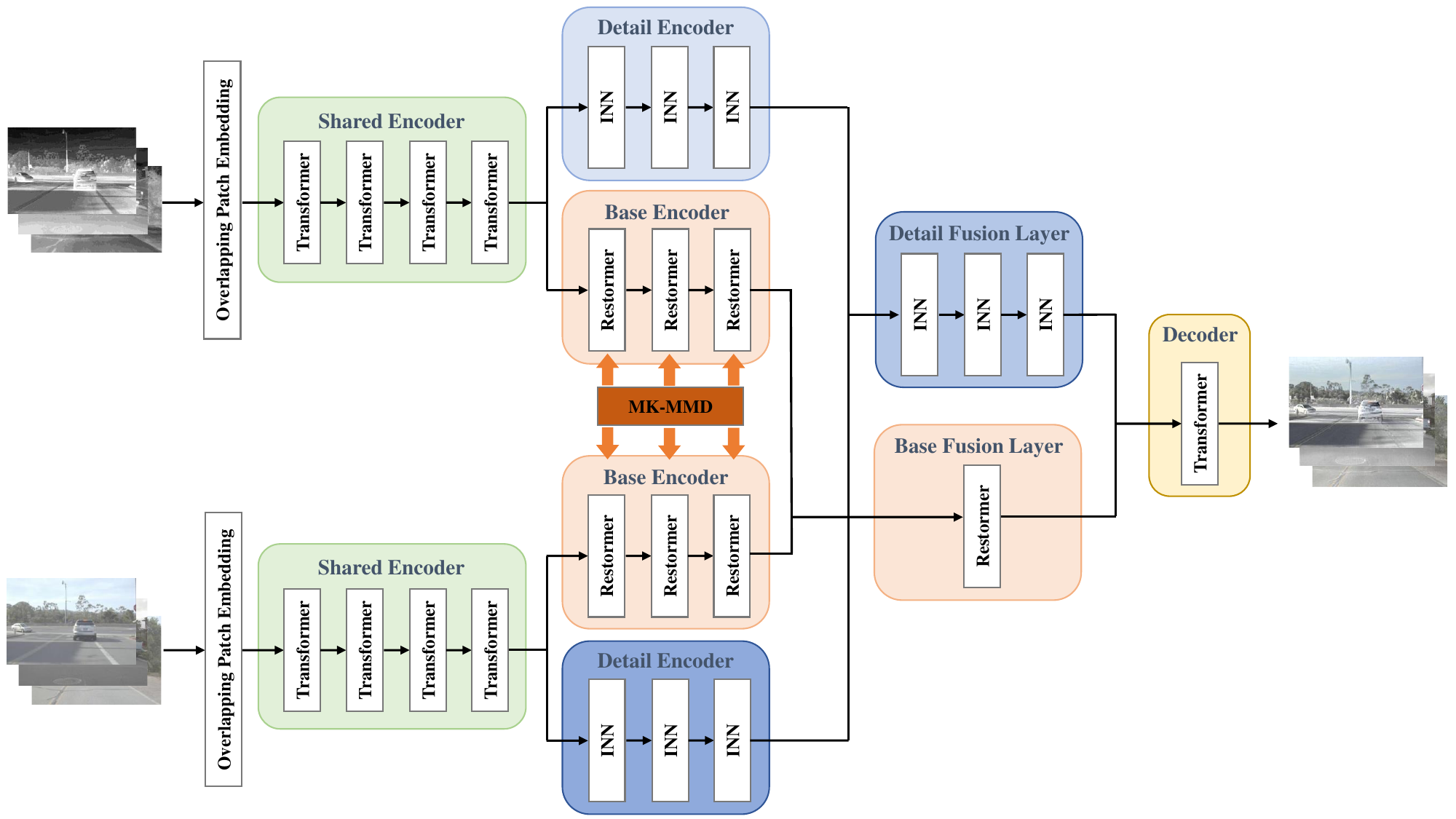}
	\caption{The framwork of the proposed DAF-Net.}
	\label{teacher_st_model}
\end{figure}

The DAF-Net consists of an encoder-decoder branch and a domain-adaptive layer based on a hybrid kernel function, as shown in Figure~\ref{teacher_st_model}. To optimize the network parameters at each training stage, a novel loss function incorporating domain adaptive loss is introduced.

\subsubsection{The encoder-decoder branches}
The encoder consists of three parts: a shared feature layer based on the Transformer, a base encoder using Restormer blocks, and a detail encoder built with INN blocks. The base encoder captures global structural information, while the detail encoder extracts fine textures. Given the input infrared and visible images, denoted as $I \in \mathbb{R}^{H \times W}$ and $V \in \mathbb{R}^{H \times W \times 3}$, the features extracted by the shared feature layer are represented as
\begin{equation}
	Y_I^S=\mathrm{E}_\mathrm{S}(I), Y_V^S=\mathrm{E}_\mathrm{S}(V),
\end{equation}
where $\mathrm{E}_\mathrm{S}(\cdot)$ represents the shared encoder. The feature extraction process for the base and detail encoders is as follows
\begin{equation}
	\begin{aligned}
		 & Y_I^B=\mathrm{E}_\mathrm{B}\left(Y_I^S\right), Y_V^B=\mathrm{E}_\mathrm{B}\left(Y_V^S\right), \\
		 & Y_I^D=\mathrm{E}_\mathrm{D}\left(Y_I^S\right), Y_V^D=\mathrm{E}_\mathrm{D}\left(Y_V^S\right).
	\end{aligned}
\end{equation}

Here, $\mathrm{E}_\mathrm{B}(\cdot)$ and $\mathrm{E}_\mathrm{D}(\cdot)$ represent the base and detail encoders. The fusion layer includes the Base Fusion and Detail Fusion layers, represented as
\begin{equation}
	Y_{IV}^B=\mathrm{F}_\mathrm{B}\left(Y_I^B, Y_V^B\right), Y_{IV}^D=\mathrm{F}_\mathrm{D}\left(Y_I^D, Y_V^D\right),
\end{equation}
where $\mathrm{F}_\mathrm{B}(\cdot)$ and $\mathrm{F}_\mathrm{D}(\cdot)$ represent the base and detail fusion layers. The decoder generates reconstructed images $\hat{I}$ and $\hat{V}$, or the fused image $\hat{F}_{IV}$
\begin{equation}
	\begin{aligned}
		\text { Stage I: }  & \hat{I}=\mathrm{D}\left(Y_I^B, Y_I^D\right), \hat{V}=\mathrm{D}\left(Y_V^B, Y_V^D\right), \\
		\text { Stage II: } & \hat{F}_{IV}=\mathrm{D}\left(Y_{IV}^B, Y_{IV}^D\right),
	\end{aligned}
\end{equation}
where $\mathrm{D}(\cdot)$ represents the decoder, using Transformer blocks as basic units.

\subsubsection{The domain adaptive layer}
The domain adaptive layer reduces the distribution discrepancy between infrared and visible light image features by computing the MK-MMD, enabling cross-modal transfer. The core idea is to align features by minimizing the distribution difference in a shared feature space. Unlike traditional methods that rely on fully connected layers, image fusion, as a regression task, requires capturing complex nonlinear relationships. Therefore, we assess the distribution discrepancy in convolutional layers, as they retain more spatial information. To address the issue of domain differences affecting feature transfer in standard encoder-decoder architectures, we introduce domain adaptive layers in the last three convolutional layers of the base encoder to align global features, while the detail encoder avoids using MK-MMD to preserve local details. By mapping images to the Reproducing Kernel Hilbert Space (RKHS) and using hybrid kernel functions to compute distribution discrepancies, image fusion performance in complex scenarios is improved.

The traditional MK-MMD employs a multi-scale Gaussian kernel, which is a linear combination of Gaussian kernels with different bandwidth parameters $\sigma$, defined as follows
\begin{equation}
	k_{\mathrm{G}}(x_{\mathrm{I}}^i, x_{\mathrm{V}}^i)=\sum_{j=1}^{K_1} \alpha_j \exp \left(-\frac{\|x_{\mathrm{I}}^i-x_{\mathrm{V}}^i\|^2}{2 \tau_j^2}\right),
\end{equation}
where $x_{\mathrm{I}}^i$ and $x_{\mathrm{V}}^i$ are the $i$-th samples from infrared and visible images, $\tau_j$ is the bandwidth of the $j$-th Gaussian kernel, controlled by the hyperparameter $\gamma$ as $\tau = 1/{\sqrt{2 \gamma}}$, $\alpha_j$ is the weight of the j-th kernel (the weights are usually non-negative,
with a sum of 1), and $K_1$ is the number of Gaussian kernels. Unlike the Gaussian kernel, the Laplacian kernel is more sensitive to edges, which is defined as
\begin{equation}
	k_{\mathrm{L}}(x_{\mathrm{I}}^i, x_{\mathrm{V}}^i)=\sum_{j=1}^{K_2} \beta_j \exp \left(-\frac{\|x_{\mathrm{I}}^i-x_{\mathrm{V}}^i\|}{\tau_j}\right).
\end{equation}

Here, $\beta_j$ is the weight of the j-th kernel (the weights are usually non-negative,
with a sum of 1). To capture both global and local details, this study combines the Gaussian and Laplacian kernels. The hybrid kernel is defined as
\begin{equation}
	k_{\mathrm{H}}(x_{\mathrm{I}}^i, x_{\mathrm{V}}^i) = c_1 k_{\mathrm{G}}(x_{\mathrm{I}}^i, x_{\mathrm{V}}^i) + c_2 k_{\mathrm{L}}(x_{\mathrm{I}}^i, x_{\mathrm{V}}^i),
\end{equation}
where $c_1$ and $c_2$ are the weights of the Gaussian and Laplacian kernels, with their sum equal to 1.
In this study, the values of $K_1$ and $K_2$ were set to 5 and 3, respectively. The parameter $\gamma$ in the Laplacian kernels was set to 0.1, 1, and 5 to vary the bandwidth. The hybrid kernel captures both global structures and local detail differences between infrared and visible images.

Our goal is to map the infrared feature $F_{\mathrm{I}}$ and the visible feature $F_{\mathrm{V}}$ into the RKHS and evaluate their distribution distance using MK-MMD
\begin{equation}
	d_{k_{\mathrm{H}}}\left(S_{\mathrm{I}}, S_{\mathrm{V}}\right)=\lVert \mathbb{E}_{x_{\mathrm{I}}^i}[F_{\mathrm{I}}] - \mathbb{E}_{x_{\mathrm{V}}^i}[F_{\mathrm{V}}] \rVert_{\mathcal{H}_k}^2,
\end{equation}
where $\mathbb{E}[\cdot]$ denotes the expectation, and $\|\cdot\|_{\mathcal{H}_k}^2$ is the squared norm in RKHS.

\subsection{Two-stage training}
A key challenge in fusing infrared and visible images is the lack of ground truth, which makes supervised learning methods ineffective. Therefore, we use a two-stage learning scheme to train DAF-Net.

\begin{figure}[htb]
	\centering
	\includegraphics[width=8.5cm]{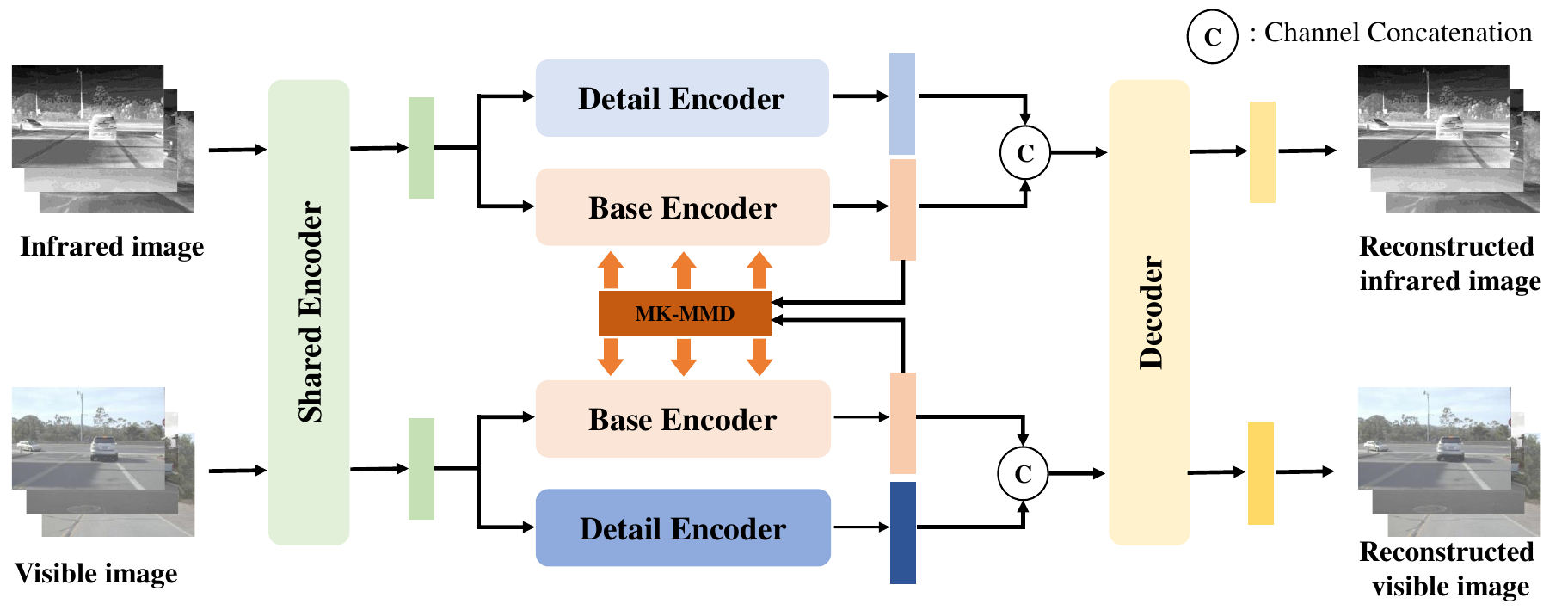}
	\caption{Codec branch training process, which there is a domain adaptive layer between the basic encoders of infrared images and visible images.}
	\label{stage_one}
\end{figure}

\subsubsection{Stage I (Encoder-decoder branches training)}
As shown in Figure~\ref{stage_one}, during training stage I, the paired infrared and visible images $I, V$ are input into a shared encoder to extract shallow features $Y_I^S, Y_V^S$. The base encoder (Restormer blocks) and detail encoder (INN blocks) then extract structural features $Y_I^B, Y_V^B$ and detail features $Y_I^D, Y_V^D$. The domain adaptive layer computes MK-MMD for the structural features. Finally, the base and detail features of the infrared (or visible) images, $Y_I^B, Y_I^D$ (or $Y_V^B, Y_V^D$), are concatenated and fed into the decoder to reconstruct $\hat{I}$ (or $\hat{V}$).

\begin{figure}[htb]
	\centering
	\includegraphics[width=8.5cm]{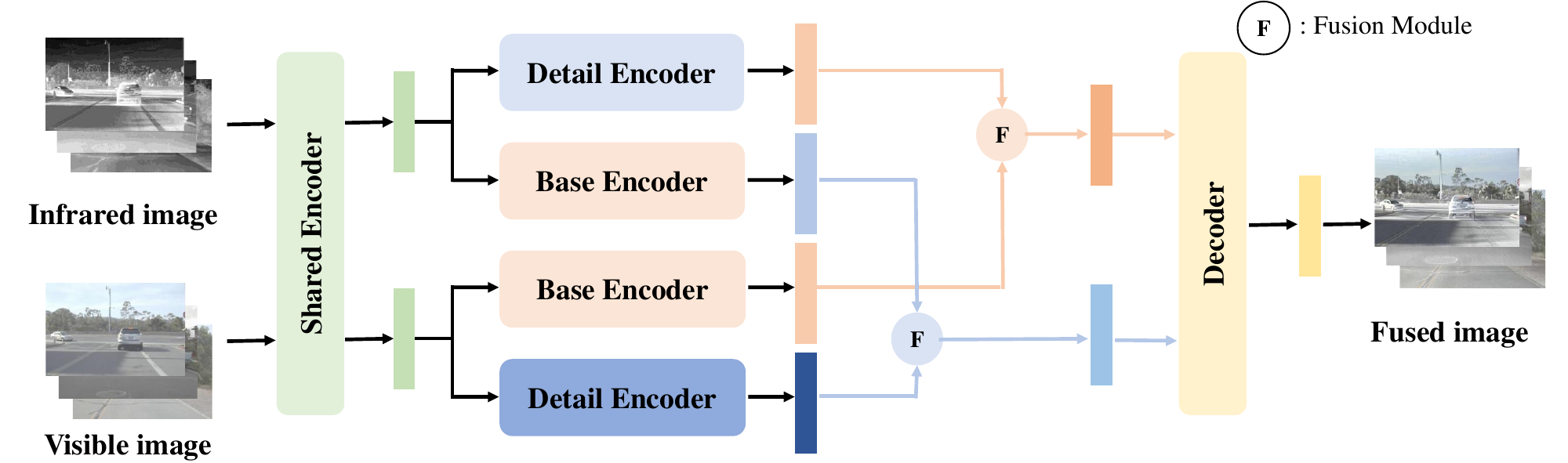}
	\caption{The process of fusing layer training.}
	\label{stage_two}
\end{figure}

\subsubsection{Stage II (Fusing layer training)}
As shown in Figure~\ref{stage_two}, during training stage II, the paired infrared and visible images $I, V$ are input into the trained encoder to obtain decomposed features. The base features $Y_I^B, Y_V^B$ and detail features $Y_I^D, Y_V^D$ are fed into the fusion layers $\mathrm{F}_\mathrm{B}$ and $\mathrm{F}_\mathrm{D}$ for structural and detail feature fusion, respectively. Finally, the fused features $Y_{IV}^B, Y_{IV}^D$ are input into the decoder to generate the fused image $\hat{F}_{IV}$.

\subsection{Loss Function}
In our training process, the loss function is divided into two stages: the encoder-decoder training stage and the fusion layer training stage. Overall, the loss function of DAF-Net is the sum of the encoder-decoder loss $\mathcal{L}_{\textit{ed}}$ and the fusion layer loss $\mathcal{L}_{\textit{fuse}}$, as follows
\begin{equation}
	\mathcal{L}_{\text{total}} = \mathcal{L}_{\text{ed}} + \mathcal{L}_{\text{fuse}}.
\end{equation}
\subsubsection{Stage I (Encoder-decoder branches training)}
During the encoder-decoder training phase, the reconstruction loss function comprises Mean Squared Error (MSE) loss, Structural Similarity Index Measurement (SSIM)\cite{SSIM} loss, and gradient loss, which is defined as follows
\begin{equation}
	\mathcal{L}_{\text{recon}} = \mathcal{L}_{\text{mse}} + \alpha_1 \mathcal{L}_{\text{ssim}} + \alpha_2 \mathcal{L}_{\text{grad}},
	\label{loss:1}
\end{equation}
where
\begin{equation}
	\begin{aligned}
		\mathcal{L}_{\text{mse}}  & = \sum_{i=1}^{N} \left( \left( V_i - \hat{V}_i \right)^2 + \left( I_i - \hat{I}_i \right)^2 \right),                                                                                         \\
		\mathcal{L}_{\text{ssim}} & = \sum_{i=1}^{N} \left(
		\frac{(2 \mu_{V_i} \mu_{\hat{V}_i} + c_1)(2 \sigma_{V_i \hat{V}_i} + c_2)}{(\mu_{V_i}^2 + \mu_{\hat{V}_i}^2 + c_1)(\sigma_{V_i}^2 + \sigma_{\hat{V}_i}^2 + c_2)} \right.                                                 \\
		                          & \phantom{=}\left. + \frac{(2 \mu_{I_i} \mu_{\hat{I}_i} + c_1)(2 \sigma_{I_i \hat{I}_i} + c_2)}{(\mu_{I_i}^2 + \mu_{\hat{I}_i}^2 + c_1)(\sigma_{I_i}^2 + \sigma_{\hat{I}_i}^2 + c_2)}\right), \\
		\mathcal{L}_{\text{grad}} & = \sum_{i=1}^{N} \left( \left\| \nabla V_i - \nabla \hat{V}_i \right\|_1 + \left\| \nabla I_i - \nabla \hat{I}_i \right\|_1 \right).
	\end{aligned}
\end{equation}

Here, \( V_i \) and \( \hat{V}_i \) represent the original and reconstructed visible images, respectively, and \( I_i \) and \( \hat{I}_i \) represent the original and reconstructed infrared images. \( \mu_{V_i} \), \( \mu_{\hat{V}_i} \), \( \mu_{I_i} \), and \( \mu_{\hat{I}_i} \) denote the mean values of visible and infrared images, while \( \sigma_{V_i}^2 \), \( \sigma_{\hat{V}_i}^2 \), \( \sigma_{I_i}^2 \), and \( \sigma_{\hat{I}_i}^2 \) represent their variances. \( \sigma_{V_i \hat{V}_i} \) and \( \sigma_{I_i \hat{I}_i} \) are the covariances. \( c_1 \) and \( c_2 \) are constants introduced to stabilize the division in the SSIM formula. Finally, \( \nabla \) represents the Sobel gradient operator used to compute the image gradients.

To capture cross-modal relationships, we introduce the correlation loss $\mathcal{L}_{\text{corr}}$, which measures the correlation between structural and detailed features, as shown below
\begin{equation}
	\mathcal{L}_{\text{corr}} = \mathcal{C}(Y_V^B, Y_I^B) + \mathcal{C}(Y_V^D, Y_I^D)
\end{equation}
where \( \mathcal{C}(\cdot) \) is the correlation coefficient operator\cite{asuero2006correlation}.

Information Noise-Contrastive Estimation (InfoNCE) loss [20] is also used during training to help model learns semantically meaningful features by contrasting positive sample pairs (from the same class) and negative sample pairs (from different classes). It is defined as:
\begin{equation}
	\mathcal{L}_{\text{InfoNCE}} = - \frac{1}{K} \sum_{i=1}^{K} \log \frac{\exp\left(\frac{\text{sim}(x_i, y_i)}{\tau}\right)}{\sum_{j=1}^{K} \exp\left(\frac{\text{sim}(x_i, y_j)}{\tau}\right)},
\end{equation}
where \( K \) denotes the batch size, \( \text{sim}(x_i, y_j) \) represents the similarity score (the dot product used here) between feature vectors \( x_i \) and \( y_j \). \( \tau \) is the temperature parameter that scales the similarity scores, set to 0.1 in this context. The loss function encourages the feature vectors of positive pairs to be similar, while pushing negative pairs farther apart, thus learning better feature representations.

To align the feature distributions of different modalities, we use a constructed hybrid kernel to calculate the distribution difference between the low-frequency features of infrared and visible images and compute the MK-MMD loss as follows
\begin{equation}
	\mathcal{L}_{\text{mkmmd}} = d_{k_{\mathrm{H}}}\left(Y_I^B, Y_V^B\right),
\end{equation}
Therefore, the loss function during the encoder-decoder training phase can be expressed as follows
\begin{equation}
	\mathcal{L}_{\text{ed}} = \mathcal{L}_{\text{recon}} + \beta_1\mathcal{L}_{\text{corr}} + \beta_2\mathcal{L}_{\text{mkmmd}} + \beta_3\mathcal{L}_{\text{InfoNCE}},
	\label{loss:2}
\end{equation}
where the weight parameters $\beta_1$, $\beta_2$, and $\beta_3$ are obtained through cross-validation.

\subsubsection{Stage II (Fusing layer training)}
During the fusion layer training phase, the loss function $\mathcal{L}_{\text{fuse}}$ consists of intensity loss, maximum gradient loss, and correlation loss, as follows
\begin{equation}
	\mathcal{L}_{\text{fuse}} = \mathcal{L}_{\text{in}} + \gamma_1 \mathcal{L}_{\text{max\_grad}} + \gamma_2 \mathcal{L}_{\text{corr}},
	\label{loss:3}
\end{equation}
where
\begin{equation}
	\begin{aligned}
		\mathcal{L}_{\text{in}}        & = \frac{1}{L} \sum_{i=1}^{L} \left\| \max(Y_i, I_i) - \hat{I}_i \right\|_1,                      \\
		\mathcal{L}_{\text{max\_grad}} & = \frac{1}{L} \sum_{i=1}^{L} \left\| \max(\nabla Y_i, \nabla I_i) - \nabla \hat{I}_i \right\|_1.
	\end{aligned}
\end{equation}

Here, \( L \) represents the pixels of the image. The weight parameters $\gamma_1$ and $\gamma_2$ are obtained through cross-validation. The intensity loss and gradient loss are used to measure the differences in intensity and gradient between the input images and the fusion result.

\section{EXPERIMENTS AND RESULTS}
\subsection{Experimental setup}
The model in this paper is trained on the MSRS \cite{MRSR} dataset (1083 pairs), RoadScene \cite{RoadScene} dataset (50 pairs), and TNO \cite{TNO} dataset (361 pairs). Part of the MSRS dataset (1083 pairs) is used for training, with the remaining portion (361 pairs) and the TNO (50 pairs) and RoadScene (25 pairs) datasets reserved for evaluation. Fusion quality is measured using metrics including Mutual Information (MI), Visual Information Fidelity (VIF), Entropy (EN), Standard Deviation (SD), Spatial Frequency (SF), edge information Q$^{\text{AB/F}}$, and Structural Similarity Index Measure (SSIM), where higher values indicate better performance. Details of these metrics are provided in \cite{MA2019153}. We evaluate our model on the Infrared-Visible Image Fusion task, comparing it to state-of-the-art methods, including unified approaches like DIF \cite{DIF} and SDNet \cite{SDNet}, and methods designed specifically for infrared and visible image fusion including TarDal \cite{Tardal}, ReCoNet \cite{ReCoNet}, RFNet \cite{RFNet}, SwinFuse \cite{Swin} and CDDFuse \cite{CDDFuse}.

\subsection{Implement details}
Experiments were conducted on a system equipped with two NVIDIA A100-SXM4-40GB GPUs. During preprocessing, the training samples were randomly cropped into \(128 \times 128\) patches. The model was trained in an unsupervised manner for 40 epochs with a batch size of 4. The Adam optimizer was employed with an initial learning rate of \(10^{-4}\), reduced by half every 10 epochs. Each transformer block contained 8 attention heads and 64 dimensions. For the loss functions in Eqs. \eqref{loss:1}, \eqref{loss:2}, and \eqref{loss:3}, the coefficients \(\alpha_1\) and \(\alpha_2\) were set to 5, \(\beta_1\) to \(\beta_3\) were assigned values of 2, 1, and 0.1, respectively, while \(\gamma_1\) and \(\gamma_2\) were set to 10 and 2. The loss function parameters were tuned to ensure that each term had comparable magnitudes.

\subsection{Qualitative Results}
\begin{figure}[htb]
	\centering
	\includegraphics[width=8.5cm]{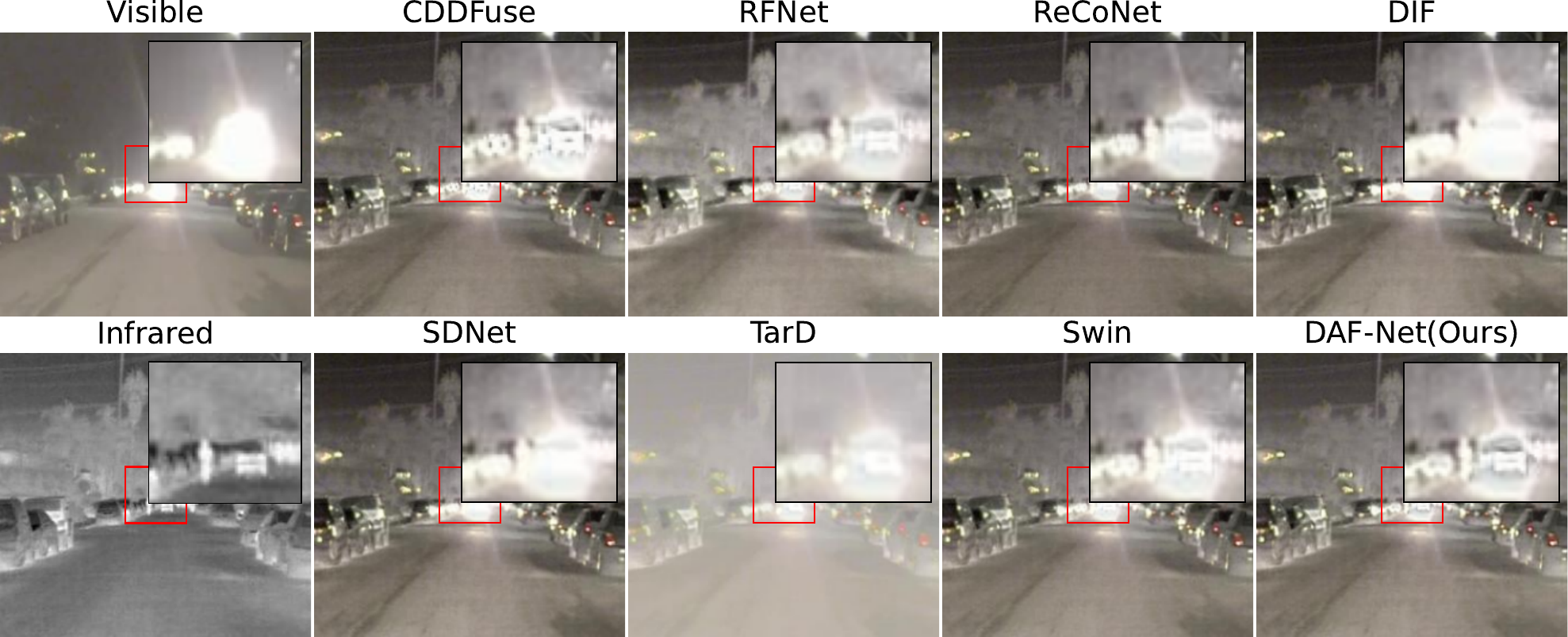}
	\caption{Comparison of results for Infrared-Visible Image Fusion task.}
	\label{qualitative}
\end{figure}

A qualitative comparison is presented in Figure~\ref{qualitative}. Obviously, our method effectively preserves the details of both infrared and visible images in areas with a lot of detail, ensuring that the details from one type of image are not overshadowed by the other. Our method effectively combines thermal radiation data from infrared images with the fine details from visible images. It enhances the visibility of objects in dark areas, making it easier to differentiate foreground targets from the background.

\subsection{Quantitative Results}

\setlength{\tabcolsep}{4pt} 
\begin{table}[htbp]
	\centering
	\caption{Dataset: TNO Infrared-Visible Image Fusion}
	\begin{tabular}{lcccccccc} 
		\hline
		Method  & EN               & SD                & SF                & MI               & SCD              & VIF              & Q$^{\text{AB/F}}$ & SSIM             \\
		\hline
		RFNet   & 6.44             & 41.16             & 11.05             & 1.87             & 1.54             & 0.69             & 0.49              & 0.55             \\
		ReCoNet & 5.76             & 39.42             & 10.74             & 1.67             & 1.32             & 0.56             & 0.47              & 0.53             \\
		DIF     & 7.10             & 43.42             & 13.12             & 2.09             & \underline{1.78} & 0.71             & 0.52              & 0.64             \\
		SDNet   & 5.72             & \underline{44.49} & \textbf{13.41}    & \underline{2.10} & 1.77             & 0.73             & 0.52              & 0.65             \\
		TarD    & 6.04             & 24.14             & 6.95              & 1.84             & 1.31             & 0.49             & 0.26              & 0.60             \\
		Swin    & 6.87             & 43.06             & 12.11             & 1.91             & 1.73             & 0.73             & 0.49              & 0.65             \\
		CDDFuse & \underline{7.11} & 45.00             & \underline{13.15} & \textbf{2.18}    & 1.76             & \underline{0.74} & \underline{0.53}  & \underline{0.66} \\
		Ours    & \textbf{7.16}    & \textbf{45.02}    & 12.63             & 2.06             & \textbf{1.80}    & \textbf{0.75}    & \textbf{0.54}     & \textbf{0.68}    \\
		\hline
	\end{tabular}
	\label{tab:sota1}
\end{table}

\setlength{\tabcolsep}{4pt} 
\begin{table}[htbp]
	\centering
	\caption{Dataset: MSRS Infrared-Visible Image Fusion}
	\begin{tabular}{lcccccccc} 
		\hline
		Method  & EN               & SD                & SF                & MI               & SCD              & VIF              & Q$^{\text{AB/F}}$ & SSIM             \\
		\hline
		RFNet   & 4.82             & 37.89             & 9.77              & 3.10             & 1.36             & 0.61             & 0.52              & 0.59             \\
		ReCoNet & 5.01             & 31.07             & 6.72              & 2.76             & 1.47             & 0.88             & 0.57              & 0.61             \\
		DIF     & 5.57             & 39.27             & 11.00             & 3.27             & 1.54             & \underline{1.01} & 0.58              & 0.66             \\
		SDNet   & 6.67             & \underline{42.46} & \underline{11.47} & \underline{3.43} & 1.55             & 0.99             & 0.65              & 0.64             \\
		TarD    & 5.03             & 32.49             & 5.13              & 2.87             & 0.99             & 0.97             & 0.59              & 0.63             \\
		Swin    & 6.55             & 42.44             & 11.40             & \underline{3.43} & \underline{1.63} & \underline{1.01} & \underline{0.67}  & 0.66             \\
		CDDFuse & \underline{6.69} & 42.37             & 11.46             & \textbf{3.47}    & 1.62             & \textbf{1.03}    & \underline{0.67}  & \underline{0.68} \\
		Ours    & \textbf{6.70}    & \textbf{43.26}    & \textbf{11.48}    & 3.13             & \textbf{1.65}    & \textbf{1.03}    & \textbf{0.68}     & \textbf{0.69}    \\
		\hline
	\end{tabular}
	\label{tab:sota3}
\end{table}

The quantitative results are shown in Table \ref{tab:sota1} and \ref{tab:sota3}. Bold indicates the best performance, and underline denotes the second-best. As observed, our method consistently outperforms others in most metrics.

\section{CONCLUSION}
This paper proposes DAF-Net with domain adaptive, using MK-MMD in the base encoder for global feature alignment while preserving modality-specific details. Experiments show superior fusion performance and applicability across datasets.

\bibliographystyle{./DAF-Net}
\bibliography{./DAF-Net}
\end{document}